\documentclass[letterpaper]{article}
\usepackage{aaai2027}
\usepackage[hyphens]{url}
\usepackage{graphicx}
\usepackage{natbib}
\usepackage{caption}
\usepackage{amsmath}
\usepackage{booktabs}
\usepackage{multirow}

\nocopyright

\title{SPT: Skills as Pre-Training Data for Agentic Language Models}
\author{
Yufei Sun\textsuperscript{\rm 1}\equalcontrib,
Yudong Li\textsuperscript{\rm 2}\equalcontrib\corresponding,
Yiming Cheng\textsuperscript{\rm 2}
}
\affiliations{
\textsuperscript{\rm 1}Beijing University of Posts and Telecommunications\\
\textsuperscript{\rm 2}Tsinghua University\\
\texttt{liyudong@tsinghua.edu.cn}
}

\begin{document}
\maketitle
\begin{abstract}
Agentic (tool-using) language models are mainly trained on tool-call traces and agent trajectories during post-training. These data provide direct behavioral supervision, but producing them requires task environments, execution, and verification, making broad tool and task coverage expensive. Publicly available skills offer another source of training data: they encode reusable tool semantics and workflows but are typically used only as inference-time context. We introduce Skill Pre-Training (SPT), a mid-training method that applies causal language modeling to \textsc{SkillCorpus}, a collection of public multi-file skill packages, optionally mixed with general data. To preserve relations among files within each package, we also introduce Reference Insert, a reference-aware assembly strategy that places supporting files near their mentions in the primary instruction. Experiments across multiple model scales and post-training recipes show that SPT consistently improves agentic performance over mid-training on general or trajectory data, while largely preserving general performance. Data mixture experiments show additional benefits from combining skill data with general annealing corpora. These results indicate that skill packages are a valuable data source for pre-training agentic language models.
\end{abstract}

\section{Introduction}

Large language models (LLMs) increasingly act as agents that use external tools for multi-step tasks~\cite{yao2022react,schick2023toolformer,hu2026agentic}. Tool use requires models to understand tool functions and carry out interactions over multiple steps. However, complete tool-use processes rarely appear in naturally collected corpora. Most models therefore acquire these capabilities during post-training through supervised fine-tuning (SFT) and reinforcement learning (RL)~\cite{chen2024agent,prabhakar2026apigen,qian2026toolrl,dong2025agentic}.

\begin{figure}[!t]
\centering
\includegraphics[width=\columnwidth]{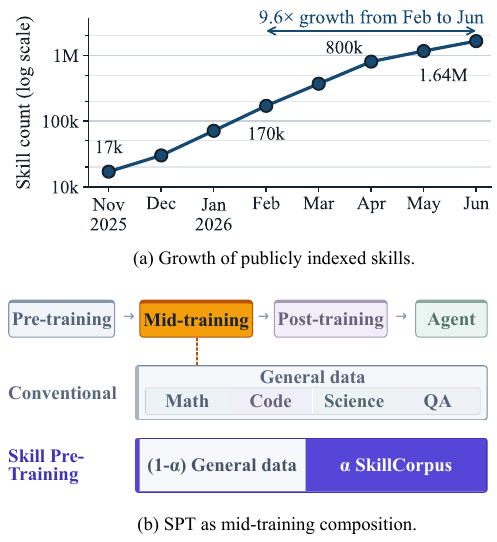}
\caption{
Public skill data growth and the SPT mid-training pipeline. (a) Snapshots from the npm registry show that the number of skills available on the platform grew $9.6\times$ from February to June 2026. (b) SPT mixes an $\alpha$ fraction of \textsc{SkillCorpus} with $(1-\alpha)$ general data during mid-training, followed by post-training.
}
\label{fig:skill_data}
\end{figure}

Existing work mainly fills this data gap with synthetic function calls and agent trajectories~\cite{tang2023toolalpaca,liu2024apigen,prabhakar2026apigen,liang2026unitoolcall,fang2025towards}. Producing valid trajectories requires task environments, execution, and verification, so the cost grows with tool and task coverage. In addition, a trajectory records one execution rather than a reusable workflow. The high production cost and execution-specific content make trajectories difficult to collect at the scale and coverage required for pre-training.

These limitations motivate a different source of agentic pre-training data: skills, which are reusable instructions written by humans for AI agents. Unlike a trajectory that records one execution, a skill describes how an agent should perform a task and can include references, scripts, templates, and configuration files. The count of publicly available skills on npm is growing rapidly. Based on snapshots from the npm registry\footnote{\url{https://www.npmjs.com/}}, the number of available skills rose from 170,226 in February 2026 to 1,640,440 in June 2026 (Figure~\ref{fig:skill_data}(a)). This scale makes skill packages a plausible source of pre-training data.

To test this hypothesis, we introduce SPT as a mid-training method that applies causal language modeling to skill data before behavior-oriented post-training. We construct \textsc{SkillCorpus} from 38,040 cleaned and decontaminated ClawHub packages\footnote{\url{https://clawhub.ai/}}. To preserve the relations between an instruction and its supporting resources, we introduce \emph{Reference Insert}, which places each referenced file near its first mention in the primary instruction file. SPT also allows \textsc{SkillCorpus} to be mixed with general data under a fixed training-token budget. The resulting checkpoint is then used to initialize downstream post-training.

We evaluate SPT across three model scales and multiple mid-/post-training configurations. Under both general instruction tuning and function-calling SFT, SPT consistently outperforms direct SFT and mid-training on general or trajectory data across all evaluated model scales, while largely preserving general performance. Data mixture experiments achieve further agentic gains when skill data is combined with general data. When SFT is followed by reinforcement learning, SPT retains its advantage over no mid-training and general-data mid-training. The consistent improvements across these experiments show that the benefit of SPT is robust to changes in model scale, mid-training data and organization, and post-training method. These results reveal the value of skill corpora as pre-training data for agentic language models.

Our contributions are as follows.
\begin{itemize}
    \item We propose multi-file skill packages as scalable agentic pre-training data and construct the 38,040-package \textsc{SkillCorpus}.

    \item We introduce SPT, which applies the causal language-modeling objective to skill data before post-training, together with Reference Insert for reference-aware package assembly.

    \item We provide controlled corpus comparisons and ablations of mixture ratio and package organization across multiple backbones and SFT settings, plus a downstream RL robustness test.
\end{itemize}

\section{Related Work}
\subsection{Training Pipelines and Agentic Mid-Training}

Modern LLM development commonly separates broad pre-training from later stages that refine the data distribution and align model behavior. Mid-training, which may include annealing, adapts a pre-trained checkpoint on curated or domain-weighted corpora, whereas post-training uses SFT, preference optimization, or RL to teach instruction following and response behavior~\cite{kaplan2020scaling,hoffmann2022training,ouyang2022training,rafailov2023direct,lambert2024tulu}. Recent open-model efforts use late-stage data selection, domain mixtures, metadata conditioning, and curated annealing corpora to strengthen targeted capabilities while retaining broad competence~\cite{gao2025metadata,li2024datacomp,team2025olmo,allal2025smollm2,liu2025instella,olmo20242}. SPT applies this mid-training framework to skill packages that encode reusable tool semantics and workflows.

Prior work has incorporated tool calls, verified function-calling examples, search traces, and interaction trajectories into language-modeling corpora; next-token prediction over such data can improve tool invocation and agentic reasoning~\cite{schick2023toolformer,liu2024apigen,wu2025masksearch,zhuang2025hephaestus}. These corpora primarily expose models to calls or realized execution paths. SPT uses the same causal language-modeling objective but changes the training unit to reusable, multi-file workflow specifications that state tool semantics, procedural constraints, and resource relations explicitly. %Its contribution therefore concerns the structure and content of the mid-training corpus, complementing rather than replacing call- and trajectory-based post-training.

Skill-centered systems instead organize skills as reusable external assets. SkillNet provides infrastructure for creating, evaluating, and connecting more than 200,000 skills through a unified ontology~\cite{liang2026skillnet}, while SkillCenter builds a source-grounded library of 216,938 skills using multi-source acquisition, quality filtering, and claim-level traceability~\cite{sha2026skillcenter}. Their primary focus is skill construction, organization, and agent-side reuse. SPT instead treats multi-file skill packages as mid-training data and studies corpus mixture and cross-file serialization under a fixed token budget.

\subsection{Agent Tuning and Evaluation}

Most prior work on tool-use specialization focuses on post-training. Function-calling corpora train API selection, schema-compliant argument generation, and call composition across large tool collections~\cite{patil2024gorilla,qin2024toolllm,tang2023toolalpaca,liang2026unitoolcall}. Agent-tuning datasets extend this supervision to multi-turn reasoning--action--observation traces and environment interactions~\cite{yao2022react,chen2024agent,song2024agentbank,prabhakar2026apigen,fang2025towards}, while RL-based methods optimize tool-use policies against task rewards. Recent studies further indicate that SFT generalization can be limited by interface-pattern memorization and that RL outcomes depend on the initial policy and reward design~\cite{chu2025sft,qian2026toolrl,dong2025agentic,gu2026agents}. SPT exposes the model to workflow-level knowledge before post-training and evaluates whether this preparation improves downstream agentic performance.

Recent benchmarks provide fine-grained diagnostics of tool-use awareness, tool selection, argument validity, function chaining, multi-step planning, and action--feedback consistency~\cite{li2023api,huang2024metatool,ye2025tooleyes,patil2025berkeley}. Broader suites additionally evaluate web interaction, operating-system control, knowledge-intensive tasks, and long-horizon behavior~\cite{qin2025aptbench,jia2025osworld,shi2026tau}. We use these benchmarks to measure how the mid-training corpus affects distinct components of agentic behavior.

\section{Skill Pre-Training}

\begin{figure}[!b]
\centering
\includegraphics[width=\columnwidth]{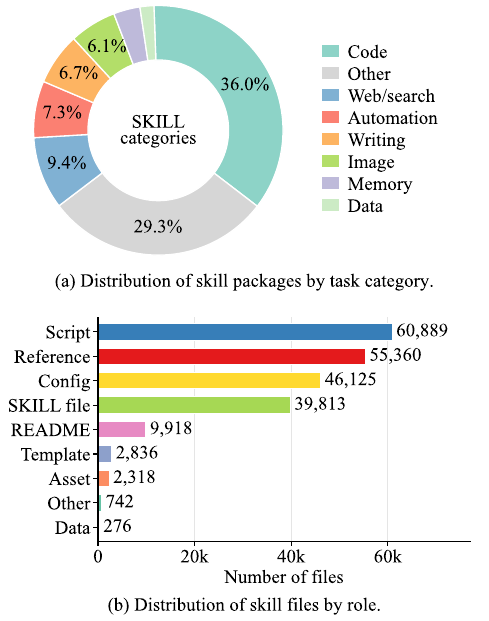}
\caption{
Task and file composition of \textsc{SkillCorpus}: (a) task categories across its 38,040 packages; (b) roles of 218,277 files.
}
\label{fig:skill_family_distribution}
\end{figure}

SPT uses human-written skill packages during mid-training, before behavior-oriented post-training. This stage exposes the base model to tool descriptions and reusable workflows before SFT teaches response and tool-call formats. Unlike a call or trajectory, which records a particular sequence of actions and outcomes, a skill explains when a capability applies, how to complete a task, which constraints to follow, and which supporting resources to use. Skills provide reusable task knowledge, whereas calls and trajectories provide direct behavioral supervision.

\begin{figure*}[!t]
\centering
\includegraphics[width=\textwidth]{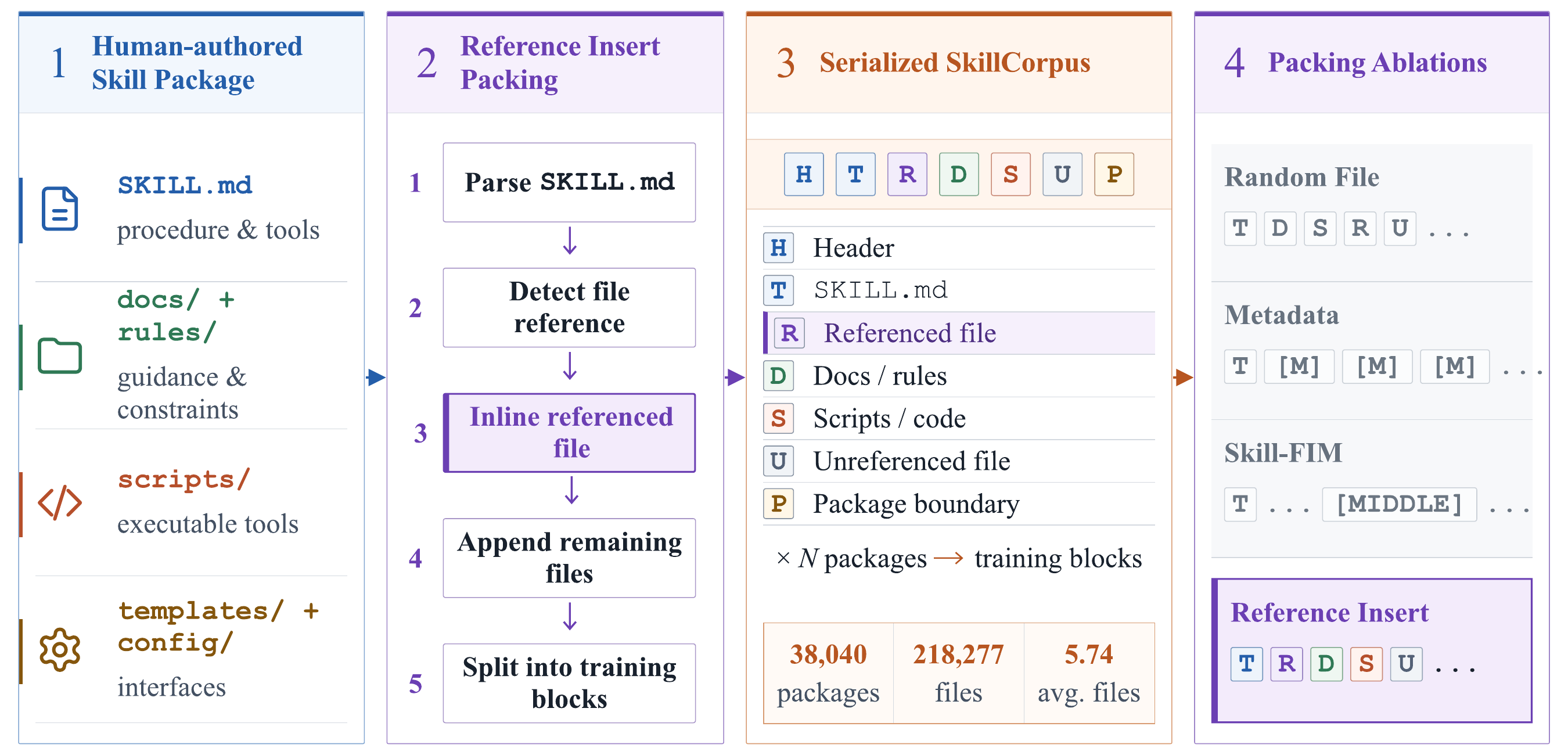}
\caption{
Overview of the SPT data pipeline.
(1) A skill package combines a primary instruction file with supporting resources; (2) Reference Insert places each referenced file near its first mention and appends the remaining files; (3) serialized packages form 4,096-token training blocks while retaining file and package boundaries; and (4) alternative file organizations are used to ablate reference-aware serialization.
}
\label{fig:spt_overview}
\end{figure*}

SPT changes the data used for mid-training without changing the standard causal language-modeling objective. It requires no task environments, tool execution, or synthetic trajectory generation. As shown in Figure~\ref{fig:spt_overview}, we collect public multi-file skill packages, clean their contents, organize each package with Reference Insert, and pack the resulting sequences into fixed-length training blocks. Reference Insert keeps instructions close to the files they mention, preserving relationships across package resources. The skill blocks can be used alone or mixed with general data during mid-training. This stage produces a skill-adapted checkpoint that initializes subsequent post-training.

\subsection{\textsc{SkillCorpus} Construction}

We construct \textsc{SkillCorpus} from a May 1, 2026 snapshot of public ClawHub packages, retaining complete packages from trusted, identity-verified, and independently audited publishers. Each package pairs a primary instruction file, typically \texttt{SKILL.md}, with optional references, scripts, templates, configurations, or other text resources; workflows may span these files.

We iteratively develop the package-level quality screen from manual audits, refining heuristics to deduplicate content, detect near-duplicate packages and autogenerated content, reject erroneous workflows, and assign quality tiers. This stage removes about 2\% of collected packages. We then remove empty or short content, encoded blobs, lock files, and binaries; redact potentially sensitive fields; and normalize paths while preserving package boundaries. Figure~\ref{fig:skill_family_distribution} summarizes the resulting task and file-role distributions; full rules and counts appear in the supplement.

For decontamination, manual audit and DeepSeek-V4-Flash screen candidates against API-Bank, MetaTool, APTBench, ToolEyes, and the Open Language Model Evaluation Standard (OLMES)~\cite{gu2025olmes}. We remove packages containing benchmark-specific tool schemas, task templates, rewritten variants, or derivatives (about 0.3\%). The final corpus contains 38,040 packages and 218,277 files (5.74 per package); the supplement gives screening criteria and overlap diagnostics.

\subsection{Multi-File Skill Assembly}

The purpose of assembly is to convert a multi-file skill package into a causal language-modeling sequence without discarding the relations expressed by file references. A simple concatenation can place an instruction far from the file it mentions. We therefore introduce Reference Insert, which makes this relation local while keeping the package's underlying files and text unchanged.

For each package, we first write a header with its name and description. We then scan \texttt{SKILL.md} in its original order and normalize candidate paths. When a line contains an unambiguous reference to a supporting file, we insert a \texttt{<referenced\_file\_sep>} marker and the referenced file immediately after that line. Each supporting file is inserted at most once, at its first unambiguous mention. After the scan, we append all remaining package files with \texttt{<unreferenced\_file\_sep>} markers. A final separator marks the boundary between packages, and the assembled stream is packed into 4,096-token training blocks. The output preserves the instruction text, supporting contents, file identities, and package boundaries in a single causal sequence. This process produces 84,905 training blocks, totaling approximately 347.8 million tokens.

We compare Reference Insert with DeepSeek-Coder Packing (DeepSeek-Coder)~\cite{guo2024deepseek,hui2024qwen2,lozhkov2024starcoder}, Random File Order (Random File), Metadata Packing (Metadata)~\cite{gao2025metadata}, and Skill fill-in-the-middle (Skill-FIM)~\cite{bavarian2022efficient}. These alternatives vary how the same package contents are organized. The ablation therefore tests whether placing referenced resources near their mentions improves the training value of skill data, rather than assuming that the effect follows from structure alone.

\subsection{Skill and General Data Mixture}

Skill data is targeted toward tool use and workflows, whereas general pre-training corpora provide broader language and knowledge coverage. SPT allows these sources to be combined under the same token budget. Let $p_{\mathrm{skill}}$ and $p_{\mathrm{gen}}$ denote the empirical distributions over assembled skill blocks and general-data blocks. Given a fixed mid-training token budget $B$, we construct the training corpus $D_{\alpha}$ as

\begin{equation}
\begin{aligned}
p_{\alpha}(x) &= \alpha p_{\mathrm{skill}}(x)
 +(1-\alpha)p_{\mathrm{gen}}(x), \\
D_{\alpha} &\sim p_{\alpha}, \qquad
|D_{\alpha}|_{\mathrm{tok}}=B, \qquad 0\leq\alpha\leq1.
\end{aligned}
\end{equation}

The coefficient $\alpha$ is the fraction of skill data. The setting $\alpha=1$ gives pure-skill SPT, while $0<\alpha<1$ mixes skill and general data. The endpoint $\alpha=0$ contains no skill data and serves only as a general-data mid-training control. We continue causal language modeling from the base model on $D_{\alpha}$; no new training objective is introduced.

We evaluate both pure-skill training and mixtures with Dolmino~\cite{olmo20242} or SmolLM~\cite{allal2025smollm2}. The pure-skill setting isolates whether skill packages are useful pre-training data relative to general and trajectory controls. The mixture sweep asks whether skills should replace general annealing data or complement it. We do not assume that one mixture is universally optimal; the experiments measure the tradeoff between agentic and general performance across the evaluated ratios.

\begin{table*}[!t]
\centering
\small
\setlength{\tabcolsep}{2.6pt}
\begin{tabular*}{\textwidth}{@{\extracolsep{\fill}}ll*{12}{c}@{}}
\toprule
\multicolumn{2}{c}{\textbf{Training Data}} &
\multicolumn{5}{c}{\textbf{Agentic Benchmarks} $\uparrow$} &
\multicolumn{7}{c}{\textbf{General Benchmarks} $\uparrow$} \\
\cmidrule(lr){1-2}\cmidrule(lr){3-7}\cmidrule(lr){8-14}
\textbf{Mid-train} & \textbf{Post-train} &
\textbf{API} & \textbf{Meta} & \textbf{APT} & \textbf{Eyes} & \textbf{Avg.} &
\textbf{ARC} & \textbf{BQ} & \textbf{HS} & \textbf{PIQA} & \textbf{WG} & \textbf{MMLU} & \textbf{Avg.} \\
\midrule
\multicolumn{14}{c}{\textbf{MeCo-1.6B-DCLM-160B}} \\
\midrule
No mid-training & \multirow{4}{*}{xLAM-FC} & 9.82 & 12.50 & 10.22 & 22.87 & 13.85 & \underline{43.94} & \underline{69.70} & \textbf{67.40} & 73.10 & \underline{64.48} & \underline{36.43} & \textbf{59.18} \\
Dolmino  &  & 15.76 & 16.16 & 17.47 & 23.43 & 18.20 & \textbf{44.37} & 69.50 & 66.30 & 72.90 & 63.85 & \textbf{36.51} & 58.91 \\
AgentBank &  & \underline{18.39} & \underline{17.13} & \underline{19.47} & \underline{27.49} & \underline{20.62} & 43.09 & 69.10 & \underline{66.80} & \textbf{73.90} & \textbf{64.64} & 36.38 & 58.99 \\
\textsc{SkillCorpus} & & \textbf{29.13} & \textbf{21.30} & \textbf{23.02} & \textbf{28.14} & \textbf{25.40} & 43.69 & \textbf{70.90} & 65.70 & \underline{73.40} & 64.09 & 36.20 & \underline{59.00} \\
\midrule
No mid-training & \multirow{4}{*}{Tulu 3} & 8.05 & 9.72 & 12.99 & 21.01 & 12.94 & 43.37 & 66.10 & \underline{67.80} & \underline{73.90} & 63.48 & \underline{36.45} & 58.52 \\
Dolmino & & 11.34 & 15.85 & 13.58 & 21.47 & 15.56 & \textbf{44.26} & \underline{67.90} & \textbf{68.10} & \textbf{75.00} & 63.19 & \textbf{37.32} & \textbf{59.29} \\
AgentBank & & \underline{14.37} & \underline{16.06} & \underline{14.51} & \textbf{26.96} & \underline{17.98} & \underline{44.20} & \textbf{68.70} & 67.70 & 73.80 & \textbf{64.17} & 36.25 & \underline{59.14} \\
\textsc{SkillCorpus} & & \textbf{26.47} & \textbf{17.39} & \textbf{17.97} & \underline{26.39} & \textbf{22.06} & 43.94 & 65.30 & 66.20 & 73.80 & \underline{63.69} & 36.24 & 58.20 \\
\midrule
\multicolumn{14}{c}{\textbf{Instella-3B-Stage1}} \\
\midrule
No mid-training & \multirow{4}{*}{xLAM-FC} & 20.77 & 17.85 & 14.72 & 32.40 & 21.44 & \underline{66.69} & \textbf{84.40} & 77.00 & \textbf{79.10} & 72.45 & \underline{54.06} & \textbf{72.28} \\
Dolmino & & 31.42 & 18.02 & 17.60 & 36.09 & 25.78 & \textbf{66.89} & 83.40 & 76.10 & 78.10 & \textbf{73.56} & \textbf{54.22} & \underline{72.05} \\
AgentBank &  & \underline{33.54} & \underline{22.46} & \underline{19.72} & \underline{38.37} & \underline{28.52} & 60.75 & 82.80 & \underline{77.10} & \underline{78.50} & \underline{72.77} & 50.10 & 70.34 \\
\textsc{SkillCorpus} & & \textbf{45.34} & \textbf{31.02} & \textbf{23.97} & \textbf{53.40} & \textbf{38.43} & 64.68 & \underline{83.70} & \textbf{78.10} & 78.20 & 72.30 & 51.62 & 71.43 \\
\midrule
No mid-training & \multirow{4}{*}{Tulu 3} & 21.87 & 11.63 & 16.64 & 31.95 & 20.52 & \underline{64.40} & 81.00 & \underline{76.20} & \textbf{77.90} & 70.30 & 52.30 & \underline{70.35} \\
Dolmino & & 26.38 & 16.37 & 18.23 & 35.67 & 24.16 & 62.50 & 80.80 & \textbf{76.30} & 77.20 & \underline{71.00} & \underline{52.40} & 70.03 \\
AgentBank & & \underline{30.00} & \underline{18.03} & \underline{18.98} & \underline{37.09} & \underline{26.03} & 63.40 & \underline{81.30} & 74.80 & \underline{77.40} & \textbf{72.45} & 52.03 & 70.23 \\
\textsc{SkillCorpus} & & \textbf{40.44} & \textbf{28.51} & \textbf{21.95} & \textbf{44.28} & \textbf{33.80} & \textbf{65.40} & \textbf{81.90} & 75.30 & \underline{77.40} & \underline{71.00} & \textbf{52.50} & \textbf{70.58} \\
\midrule
\multicolumn{14}{c}{\textbf{OLMo-3-1025-7B}} \\
\midrule
No mid-training & \multirow{4}{*}{xLAM-FC} & 28.15 & 20.93 & 25.27 & 39.64 & 28.50 & \textbf{75.51} & \textbf{87.40} & \underline{79.40} & \textbf{77.90} & \textbf{71.98} & 59.68 & \textbf{75.31} \\
Dolmino & & 36.45 & 22.87 & 30.07 & 43.92 & 33.33 & \textbf{75.51} & 86.90 & 78.70 & \underline{77.70} & 71.59 & \textbf{60.57} & \underline{75.16} \\
AgentBank & & \underline{48.82} & \underline{49.24} & \underline{31.80} & \underline{46.34} & \underline{44.05} & 73.29 & 86.70 & \textbf{79.70} & 77.10 & \underline{71.90} & 57.60 & 74.38 \\
\textsc{SkillCorpus} & & \textbf{53.23} & \textbf{53.10} & \textbf{39.96} & \textbf{67.54} & \textbf{53.46} & \underline{75.00} & \underline{87.20} & 78.50 & \textbf{77.90} & 71.82 & \underline{59.98} & 75.07 \\
\midrule
No mid-training & \multirow{4}{*}{Tulu 3} & 25.94 & 18.56 & 23.83 & 37.65 & 26.49 & 74.11 & \underline{86.10} & 80.20 & 78.20 & 72.09 & 58.69 & 74.90 \\
Dolmino & & 31.37 & 19.88 & 28.62 & 41.77 & 30.41 & \underline{74.74} & 85.80 & \underline{80.80} & \textbf{79.00} & \underline{72.93} & \underline{59.41} & \textbf{75.45} \\
AgentBank & & \underline{42.33} & \underline{47.76} & \underline{30.06} & \underline{43.98} & \underline{41.03} & \textbf{75.09} & 85.00 & \textbf{81.40} & 78.40 & \textbf{73.01} & 58.28 & 75.20 \\
\textsc{SkillCorpus} & & \textbf{50.30} & \textbf{50.03} & \textbf{33.20} & \textbf{49.74} & \textbf{45.82} & \textbf{75.09} & \textbf{86.20} & 80.60 & \underline{78.50} & 72.61 & \textbf{59.43} & \underline{75.41} \\
\bottomrule
\end{tabular*}%
\caption{
Main results for the mid-training--post-training pipeline. API, Meta, APT, Eyes, ARC, BQ, HS, and WG abbreviate API-Bank, MetaTool, APTBench, ToolEyes, ARC-Challenge, BoolQ, HellaSwag, and WinoGrande. The Avg. columns report the unweighted mean within each benchmark group. Best and second-best scores are shown in bold and underlined, respectively.
}
\label{tab:main_results}
\end{table*}

\section{Experiments}
\label{sec:experiments}

In our experiments, we evaluate the effect of SPT on the agentic capabilities of language models. The main experiment compares SPT with direct SFT, general-data mid-training, and agent-trajectory mid-training across three model scales. We apply both general instruction tuning and function-calling SFT to examine how the downstream post-training recipe affects the benefit of SPT. In addition, we compare skill--general data mixtures and multi-file skill assembly strategies, and test whether the advantage of SPT remains after downstream reinforcement learning.

\subsection{Experimental Setup}

\textbf{Backbones and datasets.} We select three base models at different scales: MeCo-1.6B-DCLM-160B~\cite{gao2025metadata,li2024datacomp}, Instella-3B-Stage1~\cite{liu2025instella}, and OLMo-3-1025-7B~\cite{team2025olmo}. All three are checkpoints after a single stage of general pre-training and have not undergone annealing or post-training, which isolates SPT from earlier adaptation stages. For mid-training, we compare Dolmino~\cite{olmo20242} as a general-data baseline, AgentBank~\cite{song2024agentbank} as an agent-trajectory baseline, and \textsc{SkillCorpus} as the skill-data condition; direct SFT without mid-training provides an additional baseline. The three mid-training conditions use the same number of training tokens. Each base or mid-training checkpoint is then trained with either Tulu 3~\cite{lambert2024tulu}, a general instruction-tuning mixture, or xLAM-FC~\cite{zhang2025xlam}, a function-calling SFT corpus.

\textbf{Training configuration.} All runs use four NVIDIA A100 80GB GPUs, BF16 precision with TF32 enabled, and gradient checkpointing. Mid-training optimizes causal language modeling for one epoch with a sequence length of 4,096, a per-device batch size of 1, 8 gradient-accumulation steps, a learning rate of $2\times10^{-5}$, weight decay of 0.1, and a cosine schedule. SFT uses assistant-only supervision for three epochs with a sequence length of 2,048, a per-device batch size of 4, 4 gradient-accumulation steps, the same learning rate and scheduler, and no weight decay. The 1.6B experiments use AdamW, while the 3B and 7B experiments use Adafactor. All reported results are means over five random seeds.

\textbf{Evaluation.} We measure agentic capability with API-Bank~\cite{li2023api}, MetaTool~\cite{huang2024metatool}, APTBench~\cite{qin2025aptbench}, and ToolEyes~\cite{ye2025tooleyes}. Together, they cover API invocation, tool selection, multi-step agentic tasks, and tool-related reasoning. We use six benchmarks from OLMES~\cite{gu2025olmes} to measure general capability: ARC-Challenge~\cite{clark2018think}, BoolQ~\cite{clark2019boolq}, HellaSwag~\cite{zellers2019hellaswag}, PIQA~\cite{bisk2020piqa}, WinoGrande~\cite{sakaguchi2021winogrande}, and MMLU~\cite{hendrycks2020measuring}. Agentic tasks use task-specific 0--5-shot prompts and are evaluated using execution-based, exact-match, or likelihood-based accuracy metrics, with unparseable outputs counted as incorrect; OLMES uses fixed 5-shot likelihood ranking. The four- and six-benchmark averages are unweighted.

\subsection{Main Results}

Across all six backbone and SFT settings in Table~\ref{tab:main_results}, \textsc{SkillCorpus} achieves the best agentic performance, outperforming both the general corpus Dolmino and the agent-trajectory corpus AgentBank. It also ranks first in 23 of the 24 individual agentic benchmark comparisons. Relative to direct SFT, SPT improves the four-benchmark agentic score by 9.11--24.96, while the six-benchmark general score changes by only $-0.85$ to $+0.51$. SPT therefore improves downstream agentic capabilities consistently across backbones and post-training recipes, with little change in general performance.

\textbf{Skill data contributes gains beyond mid-training.} Dolmino improves the four-benchmark mean over direct SFT by 2.62--4.83 across the six settings, showing that an additional language-modeling stage contributes part of the improvement. Under the same pipeline, \textsc{SkillCorpus} adds another 6.49--20.13 over Dolmino. This substantially larger margin suggests that corpus content, not merely the added training stage, drives most of the agentic gain.

\textbf{Skill data is more effective than agent trajectories for mid-training.} AgentBank consistently improves on Dolmino, showing that agent trajectories are more useful than general text for the evaluated agentic tasks. The four-benchmark score nevertheless follows the same ordering in all six blocks: Dolmino $<$ AgentBank $<$ \textsc{SkillCorpus}. Under the same training-data budget, \textsc{SkillCorpus} outperforms AgentBank by 4.07--9.91. Skills state tool applicability, constraints, and reusable procedures explicitly, whereas trajectories record the actions and observations from a particular execution. During causal language-modeling mid-training, skill data exposes reusable workflow knowledge directly instead of requiring the model to infer it from individual executions, which may improve transfer to downstream agentic tasks.

\textbf{SPT complements tool-focused SFT.} At every backbone size, the gain over direct SFT is larger after xLAM-FC than after Tulu 3: 11.55 versus 9.11 at 1.6B, 16.99 versus 13.28 at 3B, and 24.96 versus 19.33 at 7B. The larger gains with xLAM-FC suggest that tool-focused SFT makes greater use of the workflow knowledge learned during SPT. Skill descriptions and function-call supervision contribute complementary signals at different stages of training.

\subsection{Mixing Skill and General Data}

In practical pre-training pipelines, mid-training commonly uses a mixture of high-quality general data and capability-specific data. To determine how skill data should be incorporated at this stage, we vary the \textsc{SkillCorpus}-to-general-data ratio under a fixed training-token budget. This experiment provides empirical guidance for adding skill data to existing mid-training mixtures. All runs use the 1.6B backbone and Tulu 3 SFT, and we repeat the sweep with Dolmino and SmolLM as two general-data sources.

\begin{figure}[!t]
\centering
\includegraphics[width=\columnwidth]{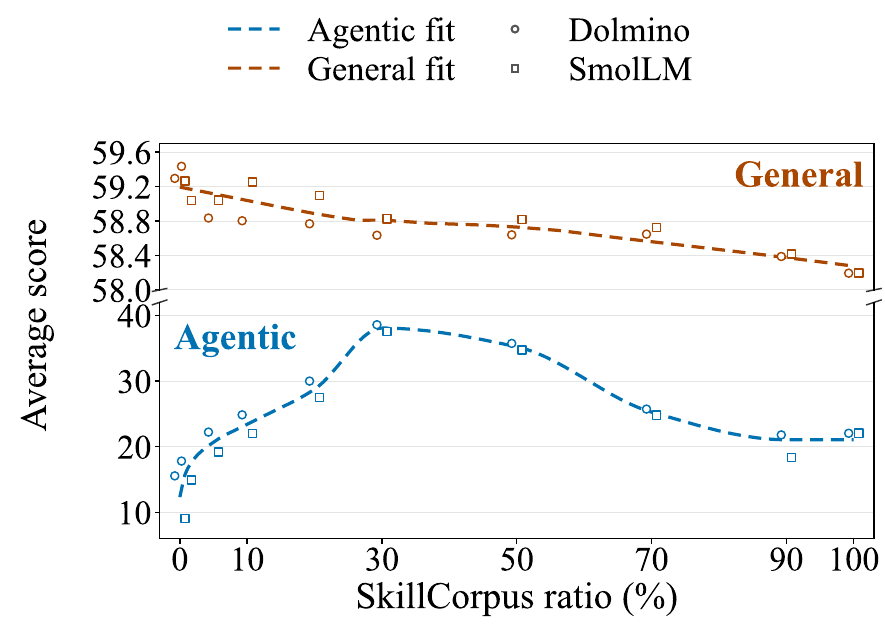}
\caption{
Agentic and general scores across \textsc{SkillCorpus} mixture ratios.
Circles denote mixtures with Dolmino, and squares denote mixtures with SmolLM; dashed curves show the fitted trends across both general-data sources.
}
\label{fig:skill_ratio_sweep}
\end{figure}

Mixed training clearly outperforms both single-source endpoints (Figure~\ref{fig:skill_ratio_sweep}). With 30\% \textsc{SkillCorpus}, the agentic score reaches 38.60 when mixed with Dolmino and 37.58 when mixed with SmolLM, compared with 15.56 and 9.07 for the corresponding general-only settings and 22.06 for pure-skill training. The same ratio gives the highest agentic score in both sweeps. Meanwhile, the general scores at this ratio remain 58.64 and 58.83, close to the general-only scores of 59.29 and 59.26. Mixing skill and general data therefore produces large agentic gains with a small change in general performance.

The benefit extends beyond a single ratio: mixtures containing 20\%--50\% skill data outperform both endpoints under both general corpora. Above 50\%, the agentic score falls rapidly, whereas the general score changes gradually. General data therefore contributes more than retention on general benchmarks; its broader language and task coverage helps the reusable workflows in skill data transfer to varied agentic tasks. Skill data is most effective as a targeted component of mid-training rather than a complete replacement for general data.

Taken together, the mixture sweeps provide a direct guideline for mid-training: skill data should be incorporated into a general-data mixture rather than used as a standalone corpus. Both the general-only and pure-skill endpoints underperform intermediate mixtures, and the same pattern holds with Dolmino and SmolLM. Joint training combines the broad language and task coverage of general data with the reusable workflows in skill data, producing the strongest downstream agentic performance.

\subsection{Multi-File Skill Assembly}

We compare our proposed Multi-File Skill Assembly strategy, Reference Insert, with existing multi-file assembly methods developed for code repositories and structured corpora. This experiment examines whether organizing files by skill-specific references improves mid-training over generic multi-file packing. We compare five strategies using the same cleaned skill files, training-token budget, tokenizer, and 4,096-token block construction. DeepSeek-Coder preserves package boundaries and original file order; Random File shuffles files within each package; Metadata adds package- and file-level attributes; Skill-FIM applies fill-in-the-middle transformations to Reference Insert sequences; and Reference Insert places supporting files after their first unambiguous mention in the primary instruction file. All models use the 1.6B backbone followed by Tulu 3 SFT.

\begin{table}[!t]
\centering
\small
\begin{tabular}{@{}l*{5}{@{\hspace{4pt}}c}@{}}
\toprule
\textbf{Strategy} & \textbf{API-Bank} & \textbf{MetaTool} & \textbf{APT} & \textbf{ToolEyes} & \textbf{Gen.} \\
\midrule
DeepSeek-Coder & \underline{24.40} & 15.22 & \underline{16.14} & 25.01 & 57.30 \\
Random File & 17.76 & 10.55 & 14.72 & 21.34 & 57.34 \\
Metadata & 11.46 & 15.03 & 13.64 & 25.86 & \underline{57.35} \\
Skill-FIM & 15.43 & \underline{17.06} & 13.37 & \underline{26.06} & 57.33 \\
Reference Insert & \textbf{26.47} & \textbf{17.39} & \textbf{17.97} & \textbf{26.39} & \textbf{58.20} \\
\bottomrule
\end{tabular}
\caption{
Effect of multi-file skill assembly on agentic and general performance.
Gen. denotes the average score of six general benchmarks.
}
\label{tab:skill_data_organization}
\end{table}

Reference Insert achieves the best result on all four agentic benchmarks and the general score. Its four-benchmark agentic score is 22.06, compared with 20.19 for DeepSeek-Coder and 16.09 for Random File. Because Reference Insert and Random File contain the same files under the same training budget, the 5.96 difference comes from how the files are arranged rather than which tokens are observed. The four alternatives obtain similar general scores of 57.30--57.35, while Reference Insert reaches 58.20, so its agentic improvement does not trade off general performance.

DeepSeek-Coder is the strongest alternative, trailing Reference Insert by 1.87. Preserving package boundaries, file paths, and source order therefore retains useful structure even without explicit reference resolution. Reference Insert further reduces the distance between an instruction and the resource it invokes, making these cross-file dependencies available within the local causal context.

Generic structural transformations do not provide the same benefit. Metadata reaches an agentic score of 16.50, while Skill-FIM reaches 17.98. Skill-FIM improves MetaTool and ToolEyes over Random File but remains below DeepSeek-Coder on API-Bank and APTBench. Adding file attributes or fill-in-the-middle transformations does not consistently connect procedural instructions with the resources they reference. The results favor reference-aware file placement over additional structural tokens or generic code-oriented packing schemes.

\subsection{SPT under Reinforcement Learning Post-Training}

Modern post-training pipelines often extend SFT with a subsequent reinforcement-learning stage. In this experiment, we test whether SPT remains effective under an SFT-then-RL pipeline. Starting from SmolLM2-360M~\cite{allal2025smollm2}, we construct three mid-training conditions: none, Dolmino, and \textsc{SkillCorpus}, with the latter two using the same training-token budget. We then apply identical Tulu 3 SFT followed by Group Relative Policy Optimization (GRPO)~\cite{shao2024deepseekmath} to all three models. GRPO uses the RLVR-GSM-MATH-IF-Mixed-Constraints dataset~\cite{lambert2024tulu}, which contains mathematical and instruction-following tasks rather than tool-use supervision.

GRPO runs for one epoch with maximum prompt and completion lengths of 1,024 and 256, four generations per prompt, a per-device batch size of 2, four gradient-accumulation steps, a learning rate of $1\times10^{-6}$, and a Kullback--Leibler (KL) coefficient of 0.04. We use Adafactor with a cosine schedule and 3\% warmup, BF16 precision, and the same five seeds.

\begin{table}[!t]
\centering
\small
\begin{tabular}{@{}l*{5}{@{\hspace{4pt}}c}@{}}
\toprule
\textbf{Mid-training} & \textbf{API-Bank} & \textbf{MetaTool} & \textbf{APT} & \textbf{ToolEyes} & \textbf{Gen.} \\
\midrule
None & 5.44 & 1.16 & 4.54 & 8.64 & 40.10 \\
Dolmino & 7.69 & 4.75 & 7.66 & 8.90 & 40.96 \\
\textsc{SkillCorpus} & \textbf{15.93} & \textbf{10.88} & \textbf{9.43} & \textbf{11.52} & \textbf{41.49} \\
\bottomrule
\end{tabular}
\caption{
Scores under identical Tulu 3 SFT and GRPO post-training for three mid-training conditions.
}
\label{tab:rlvr_results}
\end{table}

Under SFT followed by GRPO, \textsc{SkillCorpus} achieves the best result on all four agentic benchmarks (Table~\ref{tab:rlvr_results}). Its four-benchmark score is 11.94, compared with 7.25 for Dolmino and 4.95 without mid-training, giving margins of 4.69 and 6.99. It also obtains the highest general score of 41.49, compared with 40.96 and 40.10. Because all three models receive the same SFT and GRPO training, these differences reflect the effect of the preceding mid-training data.

Across the main and GRPO experiments, the results show that SPT is effective under two post-training pipelines: SFT alone and SFT followed by reinforcement learning. The RL objective does not provide tool-use supervision, yet the agentic advantage from skill mid-training remains after GRPO. SPT can therefore improve the model before different post-training methods rather than depending on a particular form of downstream optimization.

\section{Limitations and Ethical Considerations}

\textsc{SkillCorpus} is drawn from a single public repository; therefore, its domain and language coverage reflects the ClawHub community. The packages disproportionately represent certain programming languages, tool ecosystems, and task categories popular in that community, while low-resource languages and specialized agentic domains remain underrepresented. Our main experiments cover English and Chinese data, three 1.6B--7B backbones, and the reported training budgets, with a separate 360M study for RL post-training. Larger models, other compute regimes, and alternative corpus compositions remain untested. The aggregate benchmarks do not separately measure failures in tool selection, argument construction, error recovery, or unsafe action sequences, and we do not evaluate interactive long-horizon behavior.

The corpus consists of public packages from identity-verified and independently audited publishers. During corpus construction, we remove detected personal data, secrets, binaries, encoded payloads, and benchmark-derived content, although residual errors or unsafe instructions may remain. Because skills can specify executable workflows, downstream risk depends on the tools and permissions available to the model. Corpus distribution will follow source licenses and ClawHub's platform terms.

\section{Conclusion}

This work studies skill packages as a source of pre-training data, using them during mid-training before behavior-oriented post-training. We introduce SPT and evaluate its effect on downstream agentic capabilities across multiple backbones and mid-/post-training configurations. SPT consistently improves agentic performance over direct post-training and mid-training on general or trajectory data while largely preserving general performance. Mixture, assembly, and RL experiments further show that these gains remain under different corpus compositions, package organizations, and post-training methods. Skill packages can therefore provide reusable workflow knowledge during training, extending their role beyond inference-time context.

\clearpage
\appendix
\section*{Appendix}
\section{Corpus and Training Details}
\label{app:training_details}

\paragraph{Skill corpus.}
\textsc{SkillCorpus} is sourced from the public skill repository
ClawHub\footnote{\url{https://clawhub.ai/}},
where each package is a multi-file reusable workflow specification.
The collection snapshot was acquired on May 1, 2026.
We retain only packages from trusted publishers that have passed identity
verification and independent third-party audits.
The corpus spans task families including document processing, code execution,
browser operation, spreadsheet analysis, data conversion, multimodal generation,
and domain-specific tool adapters, as shown in the corpus-composition figure in
the main paper.
These task families are used only to describe the corpus; category labels are
not used during training.

\begin{center}
\small
\setlength{\tabcolsep}{3pt}
\begin{tabular}{p{0.42\columnwidth}p{0.50\columnwidth}}
\toprule
\textbf{Item} & \textbf{Value} \\
\midrule
Source collection & ClawHub skill packages \\
Collection date & May 1, 2026 \\
Training unit before serialization & Multi-file skill package \\
Primary file role & \texttt{SKILL.md}/README-style workflow specification \\
Supporting file roles & References, scripts, templates, configs, small text resources \\
Default serialization & Reference Insert with explicit package/file separators \\
Maximum package budget & 100k characters before priority-based section selection \\
\bottomrule
\end{tabular}
\captionof{table}{
Corpus provenance and serialization controls for \textsc{SkillCorpus}.
}
\label{tab:corpus_training_stats}
\end{center}

\paragraph{Mid-training corpus accounting.}
For the main comparison, Dolmino, AgentBank, and \textsc{SkillCorpus} are
materialized to the same fixed token budget. Table~\ref{tab:midtraining_token_accounting}
reports exact post-packing totals. Total training tokens include any
end-of-sequence (EOS) padding added to complete the final 4,096-token block.

\begin{center}
\footnotesize
\setlength{\tabcolsep}{4pt}
\begin{tabular}{lrr}
\toprule
\textbf{Corpus} & \textbf{Training tokens} & \textbf{Padding tokens} \\
\midrule
Dolmino & 347,770,880 & 0 \\
AgentBank & 347,770,880 & 4,065 \\
\textsc{SkillCorpus} & 347,770,880 & 351 \\
\bottomrule
\end{tabular}
\captionof{table}{
Exact token accounting for the three main mid-training corpora.
}
\label{tab:midtraining_token_accounting}
\end{center}

\paragraph{Corpus-composition proxies.}
We describe AgentBank and \textsc{SkillCorpus} using corpus-native coverage
proxies. AgentBank's 19 task configurations are grouped into four task
families and weighted by sampled non-padding tokens. The \textsc{SkillCorpus}
families use the package-level categories shown in the main-paper
corpus-composition figure. Table~\ref{tab:corpus_coverage_proxies} reports
these descriptive shares without using them as training constraints.

\begin{center}
\small
\setlength{\tabcolsep}{4pt}
\begin{tabular}{llr}
\toprule
\textbf{Corpus} & \textbf{Coverage proxy} & \textbf{Share} \\
\midrule
AgentBank & Embodied / household & 43.69\% \\
 & Web / interface / commerce & 23.77\% \\
 & Math / question answering / reasoning & 21.27\% \\
 & Code / shell / database & 11.27\% \\
\midrule
\textsc{SkillCorpus} & Code / software & 35.96\% \\
 & Other & 29.26\% \\
 & Web / search / browser & 9.39\% \\
 & Automation / workflow & 7.34\% \\
 & Document / writing & 6.74\% \\
 & Image / multimodal & 6.08\% \\
 & Memory / knowledge & 3.44\% \\
 & Data / table & 1.80\% \\
\bottomrule
\end{tabular}
\captionof{table}{
Corpus-native domain and tool-coverage proxies. AgentBank values are
sampled-token shares; \textsc{SkillCorpus} values are package shares.
}
\label{tab:corpus_coverage_proxies}
\end{center}

As a language proxy, we count basic Latin letters and Chinese, Japanese, and
Korean (CJK) unified ideographs in each materialized training text and
normalize by their combined count.
Table~\ref{tab:corpus_language_proxies} reports the resulting script shares.

\begin{center}
\small
\setlength{\tabcolsep}{7pt}
\begin{tabular}{lrr}
\toprule
\textbf{Corpus} & \textbf{Latin} & \textbf{CJK} \\
\midrule
AgentBank & 99.9991\% & 0.0009\% \\
\textsc{SkillCorpus} & 93.7496\% & 6.2504\% \\
\bottomrule
\end{tabular}
\captionof{table}{
Unicode-script proportions used as a language-composition proxy.
}
\label{tab:corpus_language_proxies}
\end{center}

\paragraph{Serialized package example.}
The abbreviated example below shows the package-level structure preserved by
the default serializer.

{\small
\begin{verbatim}
<package_sep>
ab-test-setup{cs-ab-test-setup}

<file_sep path="SKILL.md">
---
name: "ab-test-setup"
description: Plan or implement A/B tests,
  split tests, experiment variants,
  conversion experiments, or stats tests.
metadata:
  category: marketing
  updated: 2026-03-06
---

# A/B Test Setup

You are an expert in experimentation.
Your goal is to design tests that produce
valid, actionable results.

## Initial Assessment

If `.claude/product-marketing-context.md`
exists, read it before asking questions.
Before designing a test, understand:
1. Test Context -- target metric.
2. Current State -- baseline and traffic.
3. Constraints -- complexity and timeline.

## Hypothesis Framework

Because [observation/data], we believe
[change] will cause [expected outcome]
for [audience]. We will know this when
[metrics].

## Sample Size

For sample size and duration calculations:
See refs/size.md

<referenced_file_sep path="refs/size.md">
# Sample Size Guide

Reference for sample size and duration.

Required inputs:
1. Baseline conversion rate.
2. Minimum detectable effect.
3. Statistical significance level.
4. Statistical power.
\end{verbatim}
}

\paragraph{Filtering and cleaning.}
We develop the package-level quality screen through an iterative manual-audit
process. We first inspect a sample of collected packages and encode recurring
failure modes in a heuristic script. Its rules perform content-level semantic
deduplication, detect near-duplicate packages and automatically generated
content, filter erroneous workflows, and assign the remaining packages to
quality tiers. After each pass, we inspect the excluded packages, update the
rules with newly observed low-quality patterns, and rerun the screen. This
iterative process removes approximately 2\% of the collected packages.

During structural cleaning, each remaining record is parsed into a package
prefix and file sections delimited by \texttt{<file\_sep>}. We reject empty or
short records; policy-enforcing templates unrelated to reusable skills are
guarded or permission-gated.

At the file level, we discard dependency directories, lock files, binary or
media resources, archive/font/map files, files with unsupported extensions,
and long encoded blobs. In the current run, the file-level pass removes 818 lock files
and 170 encoded blobs. For retained files, the cleaner removes null bytes and
base64-encoded data uniform resource identifiers (URIs), redacts possible
secrets such as application programming interface (API) keys, tokens, passwords,
and common provider-specific key formats, and normalizes line endings,
trailing whitespace, and excessive blank lines. The cleaning pass removes 256
data URIs and applies 23,973 secret-pattern redactions.

Long code blocks, source files, and tables are kept intact during filtering.
If a package exceeds the 100k-character budget, sections are selected by priority
(\texttt{SKILL.md}, README files, other Markdown/text files,
configuration/interface files, then scripts) rather than by random truncation.
Applying this budget to 1,893 packages removes 25,632 lower-priority sections.
Packages shorter than 40 characters after cleaning are rejected.

The resulting clean package candidates preserve package boundaries, file paths,
and explicit file separators for benchmark decontamination and subsequent
serialization.

\paragraph{Training-corpus benchmark decontamination.}
Before serialization and mid-training, we screen the clean package candidates
against all public benchmarks used for evaluation. The screen covers the four
agentic benchmarks
API-Bank~\cite{li2023api}, MetaTool~\cite{huang2024metatool},
APTBench~\cite{qin2025aptbench}, and ToolEyes~\cite{ye2025tooleyes}, along
with the six Open Language Model Evaluation Standard (OLMES) benchmarks used
for general evaluation~\cite{gu2025olmes}: ARC-Challenge (ARC), BoolQ (BQ),
HellaSwag (HS), Physical Interaction: Question Answering (PIQA), WinoGrande
(WG), and Massive Multitask Language Understanding (MMLU).

Screening combines manual audit with DeepSeek-V4-Flash
(\texttt{deepseek-v4-flash}) judgments. We remove an entire skill package when
any file contains benchmark-specific tool formats or schemas, task templates,
rewritten or paraphrased variants, or other derivative content. Package-level
removal prevents supporting files from retaining related benchmark material.
Approximately 0.3\% of candidate packages are excluded from the training
corpus. After quality and structural cleaning followed by benchmark
decontamination, all \textsc{SkillCorpus} variants use the same final package
set, while the evaluation instances remain unchanged.

As a lexical diagnostic, we normalize cleaned candidates and benchmark inputs
into lowercase alphanumeric and Chinese-character tokens and compute exact
13-gram overlap. Table~\ref{tab:contamination_analysis} reports results before
package removal. The 1.24B-character candidate pool contains 163 unique
benchmark 13-grams. Matches flag candidates for review but are not the sole
removal criterion because exact overlap misses rewritten or derivative forms.

\begin{table}[!htbp]
\centering
\footnotesize
\setlength{\tabcolsep}{2.5pt}
\begin{tabular}{lrrrr}
\toprule
\textbf{Benchmark} & \textbf{Instances} & \textbf{Matched} & \textbf{13-gram Rate} & \textbf{Max Inst.} \\
\midrule
\multicolumn{5}{l}{\emph{Agentic benchmarks}} \\
API-Bank & 1,602 & 0 & 0.0000\% & 0.00\% \\
MetaTool & 6,214 & 13 & 0.0022\% & 1.76\% \\
APTBench & 3,561 & 66 & 0.0179\% & 4.65\% \\
ToolEyes & 1,121 & 40 & 0.0269\% & 4.34\% \\
\multicolumn{5}{l}{\emph{OLMES benchmarks}} \\
ARC & 2,343 & 0 & 0.0000\% & 0.00\% \\
BQ & 2,000 & 0 & 0.0000\% & 0.00\% \\
HellaSwag & 2,000 & 0 & 0.0000\% & 0.00\% \\
PIQA & 2,000 & 0 & 0.0000\% & 0.00\% \\
WG & 1,268 & 0 & 0.0000\% & 0.00\% \\
MMLU & 28,084 & 8 & 0.0014\% & 1.64\% \\
\bottomrule
\end{tabular}
\caption{
Exact-overlap diagnostic on the cleaned candidate pool before package-level
decontamination.
The \emph{Matched} column counts benchmark instances sharing at least one
normalized 13-gram with the candidate pool. The \emph{13-gram Rate} is the
fraction of unique benchmark 13-grams matched by the candidate pool; Max Inst. is the largest matched
13-gram fraction for any single benchmark instance. Benchmark instances remain
in the evaluation sets.
}
\label{tab:contamination_analysis}
\end{table}

Tables~\ref{tab:training_hyperparameters},
\ref{tab:training_hyperparameters_7b}, and
\ref{tab:training_hyperparameters_3b} list the matched training
configurations used with Tulu 3 supervised fine-tuning (SFT)~\cite{lambert2024tulu}
and xLAM function-calling (xLAM-FC) SFT~\cite{zhang2025xlam} across the MeCo
1.6B~\cite{gao2025metadata,li2024datacomp}, OLMo
7B~\cite{team2025olmo}, and Instella 3B~\cite{liu2025instella} backbones.
All runs use Brain Floating Point (BF16) precision with
TensorFloat-32 (TF32) enabled unless noted otherwise.
For every configuration, we average results over runs with seeds 42, 3407,
1234, 2026, and 2027.

\begin{table}[!htbp]
\centering
\footnotesize
\setlength{\tabcolsep}{2pt}
\begin{tabular}{p{0.75in}p{0.80in}p{0.70in}p{0.70in}}
\toprule
\textbf{Parameter} & \textbf{Mid-training} & \textbf{Tulu 3 SFT} & \textbf{xLAM-FC SFT} \\
\midrule
GPUs & 4 $\times$ A100 80GB & 4 $\times$ A100 80GB & 4 $\times$ A100 80GB \\
Backbone / initialization & MeCo-1.6B-DCLM-160B & Mid-training checkpoint; the no-mid-training condition uses the backbone directly & Mid-training checkpoint; the no-mid-training condition uses the backbone directly \\
Training objective & Causal language modeling on fixed token blocks & Assistant-only supervised chat loss & Assistant-only function-call loss \\
Training data & Mid-training corpus (experiment-specific) & Tulu 3 & xLAM-FC \\
Total budget & 1 epoch & 3 epochs & 3 epochs \\
Sequence length & 4,096 & 2,048 & 2,048 \\
Per-device batch size & 1 & 4 & 4 \\
Gradient accum. steps & 8 & 4 & 4 \\
Learning rate & $2\times10^{-5}$ & $2\times10^{-5}$ & $2\times10^{-5}$ \\
Weight decay & 0.1 & 0.0 & 0.0 \\
Optimizer & AdamW & AdamW & AdamW \\
Scheduler & cosine & cosine & cosine \\
Precision & BF16; TF32 enabled & BF16; TF32 enabled & BF16; TF32 enabled \\
Gradient checkpointing & enabled & enabled & enabled \\
No. of seeds & 5 & 5 & 5 \\
\bottomrule
\end{tabular}
\caption{
Training hyperparameters for the 1.6B experiments with mid-training followed by post-training.
Both Tulu 3 and xLAM-FC SFT settings are used for the main mid-training corpus
comparison; the Tulu 3 settings are additionally used for the mixture-ratio and
data-organization experiments.
}
\label{tab:training_hyperparameters}
\end{table}

\begin{table}[!htbp]
\centering
\footnotesize
\setlength{\tabcolsep}{2pt}
\begin{tabular}{p{0.72in}p{0.82in}p{0.62in}p{0.62in}}
\toprule
\textbf{Parameter} & \textbf{Mid-training} & \textbf{Tulu 3 SFT} & \textbf{xLAM-FC SFT} \\
\midrule
GPUs & 4 $\times$ A100 80GB & 4 $\times$ A100 80GB & 4 $\times$ A100 80GB \\
Backbone / initialization & OLMo-3-1025-7B stage1-step999000 & Mid-training checkpoint; the no-mid-training condition uses the backbone directly & Mid-training checkpoint; the no-mid-training condition uses the backbone directly \\
Training objective & Causal language modeling on fixed token blocks & Assistant-only supervised chat loss & Assistant-only function-call loss \\
Training data & Mid-training corpus (experiment-specific) & Tulu 3 & xLAM-FC \\
Total budget & 1 epoch & 3 epochs & 3 epochs \\
Sequence length & 4,096 & 2,048 & 2,048 \\
Per-device batch size & 1 & 4 & 4 \\
Gradient accum. steps & 8 & 4 & 4 \\
Learning rate & $2\times10^{-5}$ & $2\times10^{-5}$ & $2\times10^{-5}$ \\
Weight decay & 0.1 & 0.0 & 0.0 \\
Optimizer & Adafactor & Adafactor & Adafactor \\
Scheduler & cosine & cosine & cosine \\
Precision & BF16; TF32 enabled & BF16; TF32 enabled & BF16; TF32 enabled \\
Gradient checkpointing & enabled & enabled & enabled \\
No. of seeds & 5 & 5 & 5 \\
\bottomrule
\end{tabular}
\caption{
Training hyperparameters for the 7B OLMo experiments with mid-training followed by post-training.
}
\label{tab:training_hyperparameters_7b}
\end{table}

\begin{table}[!htbp]
\centering
\footnotesize
\setlength{\tabcolsep}{2pt}
\begin{tabular}{p{0.72in}p{0.82in}p{0.62in}p{0.62in}}
\toprule
\textbf{Parameter} & \textbf{Mid-training} & \textbf{Tulu 3 SFT} & \textbf{xLAM-FC SFT} \\
\midrule
GPUs & 4 $\times$ A100 80GB & 4 $\times$ A100 80GB & 4 $\times$ A100 80GB \\
Backbone / initialization & Instella-3B-Stage1 & Mid-training checkpoint; the no-mid-training condition uses the backbone directly & Mid-training checkpoint; the no-mid-training condition uses the backbone directly \\
Training objective & Causal language modeling on fixed token blocks & Assistant-only supervised chat loss & Assistant-only function-call loss \\
Training data & Mid-training corpus (experiment-specific) & Tulu 3 & xLAM-FC \\
Total budget & 1 epoch & 3 epochs & 3 epochs \\
Sequence length & 4,096 & 2,048 & 2,048 \\
Per-device batch size & 1 & 4 & 4 \\
Gradient accum. steps & 8 & 4 & 4 \\
Learning rate & $2\times10^{-5}$ & $2\times10^{-5}$ & $2\times10^{-5}$ \\
Weight decay & 0.1 & 0.0 & 0.0 \\
Optimizer & Adafactor & Adafactor & Adafactor \\
Scheduler & cosine & cosine & cosine \\
Precision & BF16; TF32 enabled & BF16; TF32 enabled & BF16; TF32 enabled \\
Gradient checkpointing & enabled & enabled & enabled \\
No. of seeds & 5 & 5 & 5 \\
\bottomrule
\end{tabular}
\caption{
Training hyperparameters for the Instella-3B experiments with mid-training followed by post-training.
}
\label{tab:training_hyperparameters_3b}
\end{table}

\section{Evaluation Protocols and Output Parsing}
\label{app:evaluation_protocols}

We follow the prompts, interaction procedures, and task metrics released with
each benchmark. The skill-corpus decontamination described above is applied
before mid-training, and every model is evaluated on the same instances. Scores
use a 0--100 scale. Except for the recovery procedures noted below, unparseable
outputs are scored as incorrect.

\paragraph{API-Bank.}
We use the three official evaluation levels of API-Bank~\cite{li2023api}: Level
1 supplies the relevant API description and evaluates API calling; Level 2 exposes
only the ToolSearcher interface and additionally requires API retrieval; and
Level 3 evaluates planning over multiple API calls. We use the released
execution-based evaluator.
The model is instructed to emit an API request in the form
\texttt{[ApiName(key='value', ...)]}. The parser first searches the response
for a bracketed function call, extracts the API name, and converts quoted
scalars, list literals, and unquoted word-valued arguments into a parameter
dictionary. It then invokes the predicted API through the official
ToolManager. Each API's task-specific \texttt{check\_api\_call\_correctness}
method compares the executed result with the reference result, allowing the
evaluator to apply API-specific equivalence rather than string equality. A
missing call, a parsing failure, an invalid API or parameter, an execution
exception, or a result mismatch causes the response to be scored as incorrect. We compute API-call accuracy separately for
the three levels and report
\begin{equation}
S_{\mathrm{API}}=\frac{1}{3}\sum_{\ell=1}^{3}\operatorname{Acc}_{\ell}.
\end{equation}
The official response-after-API ROUGE-L diagnostic is not included in this
aggregate.

\paragraph{MetaTool.}
MetaTool~\cite{huang2024metatool} separates tool-usage awareness from tool
selection. We evaluate only the four official tool-selection subtasks: choosing
among similar tools, choosing a tool within a specified scenario, rejecting the
available tools when the correct tool is absent, and selecting two tools for a
multi-tool request. The tool-usage-awareness task and its accuracy, precision,
recall, and F1 measures are excluded. Generation follows the released setup
with deterministic decoding.

For output matching, the parser normalizes candidate tool names and searches
for their literal normalized occurrences in the response, retaining matched
tools in response order. If no candidate name is found, it falls back to the
first text segment delimited by a newline or punctuation. A single-tool
subtask uses the first parsed label; the multi-tool subtask requires the set
of the first two parsed labels to equal the reference set. All 4,287
tool-selection outputs per model are parsed and scored automatically, and none
are manually adjudicated. For subtask $k$, Correct Selection Rate (CSR) is the
fraction of instances whose selected label or label set equals the reference.
The reported MetaTool score is
\begin{equation}
S_{\mathrm{Meta}}=\frac{1}{4}\sum_{k=1}^{4}\operatorname{CSR}_{k}.
\end{equation}

\paragraph{APTBench.}
We follow the official software-engineering (SWE) and deep-research (DR)
evaluation protocols of APTBench~\cite{qin2025aptbench}.
The SWE tasks cover environment setup and issue fixing through planning,
action, error handling, bug localization, fix-patch selection, and test-patch
selection. The DR tasks cover planning and action for closed-ended questions,
and planning, report selection, and citation for open-ended questions. All
tasks use the official 3-shot prompts, except open-ended report selection,
which uses an official 2-shot prompt. Decoding is greedy. If a prompt exceeds the model's
maximum sequence length, equal-sized portions are retained from its head and
tail, following the official truncation rule.

The official parser is applied separately by question type. For a
multiple-choice (MC) question, asterisks are removed and the first answer
letter in the expected parenthesized or line-terminated form is extracted. For
SWE text-completion actions, if a newline is present, the parser retains the
first line and truncates it at the first semicolon; otherwise, it keeps the
full response. The resulting command is compared with the reference by exact
match (EM). For a DR closed-ended answer, the parser retains the text before
the closing bracket supplied by the
prompt and computes both EM and ROUGE-1 F1. For citation questions, it retains
the text before the closing parenthesis, splits comma-separated option labels,
and requires exact set equality, so order is ignored but missing or extra
citations are incorrect. MC tasks use accuracy, while text-completion tasks use
EM; the two closed-ended DR action tasks additionally contribute their
ROUGE-1 scores, as in the official results table.

The SWE score is the unweighted mean of its eight reported task metrics. The DR
score is the unweighted mean of its eleven reported metrics, including both EM
and ROUGE-1 for the English and Chinese closed-ended action tasks. We combine
the two domain scores directly:
\begin{equation}
S_{\mathrm{APT}}=\frac{1}{2}
\left(S_{\mathrm{SWE}}+S_{\mathrm{DR}}\right).
\end{equation}

\paragraph{ToolEyes.}
ToolEyes~\cite{ye2025tooleyes} evaluates interactive tool use across seven
real-world scenarios. We use its official 5-shot Reasoning and Acting (ReAct)
prompt, a maximum of nine interaction turns, a temperature of 0.3, and a
top-$p$ value of 0.5.
Every assistant turn must contain, in order, \texttt{Thought:},
\texttt{Action:}, and \texttt{Action Input:}. The released parser extracts the
three fields with multiline regular expressions and requires
\texttt{Action Input} to be a valid dictionary. A malformed turn receives the
official format-correction observation and may be regenerated. A valid action
is executed against the scenario tool library, and its observation is appended
to the interaction history. The model must terminate with the \texttt{finish}
tool; exceeding the turn limit forces an unsuccessful termination.

We retain the five capability scores and equations defined in the ToolEyes
paper. Let $n$ be the number of assistant turns, $n_f$ the number satisfying
the required format, and let all model-judged rubric scores be on a 1--10
scale. Format alignment is $\mathrm{IF}=n_f/n$. Intent comprehension is
$\mathrm{IU}=q_{\mathrm{focus}}/10$. Behavior planning is the product of
thought validity and logical integrity:
\begin{equation}
\mathrm{BP}=\frac{q_{\mathrm{validity}}}{10}
\,\frac{q_{\mathrm{integrity}}}{10}.
\end{equation}
For each of the $m$ parseable turns, $r_{\mathrm{tool},j}$ is 1 only when the
selected tool exists and all parameters conform to its documentation, and
$q_{\mathrm{match},j}$ measures whether that tool agrees with the stated
thought. Thus,
\begin{equation}
\mathrm{TS}=\frac{1}{m}\sum_{j=1}^{m}
r_{\mathrm{tool},j}\frac{q_{\mathrm{match},j}}{10},
\end{equation}
where malformed turns are omitted and $\mathrm{TS}=0$ when no turn is
parseable. Following the paper rather than the released evaluator's simplified
implementation, we include the completion gate in the answer-organization score:
\begin{equation}
\mathrm{AO}=I_{\mathrm{finish}}\frac{q_{\mathrm{answer}}}{10},
\end{equation}
where $I_{\mathrm{finish}}=1$ only when the model completes the task within the
turn limit. The overall ToolEyes score is
\begin{equation}
S_{\mathrm{Eyes}}=\frac{1}{5}
(\mathrm{IF}+\mathrm{IU}+\mathrm{BP}+\mathrm{TS}+\mathrm{AO}).
\end{equation}
We multiply this value by 100 for reporting.
We preserve the official evaluation rubrics and replace the original GPT-4
judge with DeepSeek-V4-Flash (\texttt{deepseek-v4-flash})
for intent comprehension, behavior-planning validity and logical integrity,
thought--action agreement, and answer quality. The deterministic checks for
format, tool existence, and parameter validity remain unchanged.
A human audit of 300 judge-scored samples yielded 97.2\% agreement between
DeepSeek-V4-Flash and the human annotations.

\paragraph{OLMES general benchmarks.}
We use the OLMES protocol for ARC-Challenge, BoolQ, HellaSwag, PIQA,
WinoGrande, and MMLU~\cite{gu2025olmes}.
Each task uses its fixed set of five in-context examples. OLMES evaluates both the
original MC formulation, in which answer labels are ranked, and a
rank-classification or cloze formulation (RC), in which the answer texts
are candidate continuations. These are likelihood evaluations rather than
free-form generations, so no textual answer parser is used. The MC prediction
is the option with the largest summed conditional log-likelihood. For RC,
ARC-Challenge uses unconditional normalization,
\begin{equation}
s_i=\log p(a_i\mid x)-\log p(a_i),
\end{equation}
BoolQ and WinoGrande use the raw summed log-likelihood, and HellaSwag, PIQA, and
MMLU use log-likelihood normalized by the answer's character length. The RC
prediction is $\arg\max_i s_i$. The score for each benchmark is the higher of
its MC and RC accuracies, as specified by OLMES. MMLU first macro-averages over
its subjects. Finally, the general score is the unweighted mean of the six
benchmark scores.

\section{Full Skill Mixture Ratio Results}
\label{app:full_ratio}

Tables~\ref{tab:full_skill_mixture_agentic} and
\ref{tab:full_skill_mixture_general} give the per-benchmark results for the
mixture-ratio sweep in the main paper, where \textsc{SkillCorpus} is mixed with
Dolmino~\cite{olmo20242} or SmolLM~\cite{allal2025smollm2}. Aggregate columns
are omitted because the main figure already reports the trends in the agentic
and general averages.
In all result tables below, boldface and underlining mark the best and second-best
values within each comparison block, respectively; ties share the same style.

\begin{table}[!htbp]
\centering
\small
\setlength{\tabcolsep}{2pt}
\begin{tabular}{lcccc}
\toprule
\textbf{Ratio} & \textbf{API-Bank} & \textbf{MetaTool} & \textbf{APTBench} & \textbf{ToolEyes} \\
\midrule
\multicolumn{5}{c}{\textbf{\textsc{SkillCorpus} + Dolmino}} \\
\midrule
0/100 & 11.34 & 15.85 & 13.58 & 21.47 \\
1/99 & 15.05 & 12.46 & 13.91 & 29.84 \\
5/95 & 17.41 & 12.42 & 13.81 & 45.29 \\
10/90 & 18.34 & 16.25 & 24.02 & 40.84 \\
20/80 & 31.77 & 22.50 & 23.88 & 41.88 \\
30/70 & \textbf{45.92} & \textbf{28.22} & \textbf{33.39} & \underline{46.86} \\
50/50 & \underline{38.74} & \underline{23.87} & \underline{28.81} & \textbf{51.57} \\
70/30 & 32.07 & 20.86 & 13.38 & 36.65 \\
90/10 & 22.14 & 15.90 & 13.04 & 36.13 \\
100/0 & 26.47 & 17.39 & 17.97 & 26.39 \\
\midrule
\multicolumn{5}{c}{\textbf{\textsc{SkillCorpus} + SmolLM}} \\
\midrule
0/100 & 6.94 & 2.25 & 12.91 & 14.19 \\
1/99 & 14.60 & 14.34 & 15.77 & 15.03 \\
5/95 & 15.69 & 16.86 & 19.62 & 24.50 \\
10/90 & 17.07 & 18.76 & 22.57 & 29.53 \\
20/80 & 26.94 & 21.14 & 26.77 & 35.29 \\
30/70 & \textbf{41.93} & \textbf{26.87} & \textbf{33.62} & \underline{47.91} \\
50/50 & \underline{35.79} & \underline{21.48} & \underline{30.54} & \textbf{51.05} \\
70/30 & 24.83 & 17.00 & 23.08 & 34.29 \\
90/10 & 19.15 & 15.75 & 14.45 & 23.98 \\
100/0 & 26.47 & 17.39 & 17.97 & 26.39 \\
\bottomrule
\end{tabular}
\caption{
Agentic benchmark results for different mixture ratios of
\textsc{SkillCorpus} with Dolmino and SmolLM.
}
\label{tab:full_skill_mixture_agentic}
\end{table}

\begin{table}[!htbp]
\centering
\small
\setlength{\tabcolsep}{1mm}
\begin{tabular}{lcccccc}
\toprule
\textbf{Ratio} & \textbf{ARC} & \textbf{BQ} & \textbf{HS} & \textbf{PIQA} & \textbf{WG} & \textbf{MMLU} \\
\midrule
\multicolumn{7}{c}{\textbf{\textsc{SkillCorpus} + Dolmino}} \\
\midrule
0/100 & \underline{44.26} & 67.90 & \textbf{68.10} & \textbf{75.00} & 63.19 & \textbf{37.32} \\
1/99 & \textbf{44.53} & \underline{69.20} & \underline{67.20} & \underline{73.80} & \textbf{64.98} & \underline{36.90} \\
5/95 & 43.94 & 68.70 & \underline{67.20} & 73.00 & 63.85 & 36.32 \\
10/90 & 44.20 & 68.70 & 67.00 & 73.30 & 63.38 & 36.24 \\
20/80 & 43.94 & 68.60 & 66.70 & 73.10 & 63.85 & 36.42 \\
30/70 & 43.34 & 68.10 & 66.70 & 73.60 & 63.54 & 36.53 \\
50/50 & 43.77 & 68.30 & 66.60 & 73.10 & 63.85 & 36.22 \\
70/30 & 43.60 & 68.50 & 66.60 & 72.90 & \underline{64.01} & 36.28 \\
90/10 & 42.86 & \textbf{69.30} & 66.00 & 73.00 & 63.21 & 35.96 \\
100/0 & 43.94 & 65.30 & 66.20 & \underline{73.80} & 63.69 & 36.24 \\
\midrule
\multicolumn{7}{c}{\textbf{\textsc{SkillCorpus} + SmolLM}} \\
\midrule
0/100 & \textbf{44.71} & 69.80 & \textbf{67.10} & 73.40 & \textbf{64.09} & 36.48 \\
1/99 & 44.45 & 69.50 & \underline{67.00} & 72.80 & \underline{64.01} & 36.46 \\
5/95 & 44.37 & 69.70 & \underline{67.00} & 73.20 & 63.54 & 36.44 \\
10/90 & \underline{44.54} & \textbf{70.20} & 66.70 & \underline{73.70} & \underline{64.01} & 36.37 \\
20/80 & 44.37 & 69.60 & 66.90 & 73.50 & 63.69 & \textbf{36.52} \\
30/70 & 44.11 & 68.90 & 66.80 & 73.20 & 63.46 & \underline{36.49} \\
50/50 & 43.43 & \underline{69.90} & 66.60 & 73.20 & 63.46 & 36.32 \\
70/30 & 43.34 & 68.50 & 66.60 & 73.50 & \underline{64.01} & 36.39 \\
90/10 & 43.46 & 69.30 & 66.20 & 73.00 & 62.66 & 35.89 \\
100/0 & 43.94 & 65.30 & 66.20 & \textbf{73.80} & 63.69 & 36.24 \\
\bottomrule
\end{tabular}
\caption{
Results on general benchmarks for different skill-to-general-data mixture ratios.
}
\label{tab:full_skill_mixture_general}
\end{table}

\section{Full Skill-Aware Data Organization Results}
\label{app:full_packing}

The skill-aware data organization ablation compares five serialization
strategies. All variants start from the same cleaned skill-package collection
and use the same tokenizer, validation-split ratio, end-of-sequence-token
insertion between serialized records, and five-seed averaging protocol before
being converted into 4,096-token mid-training blocks.

DeepSeek-Coder Packing (DeepSeek-Coder)~\cite{guo2024deepseek,hui2024qwen2,lozhkov2024starcoder}
serializes each cleaned package in its original source order. Each record
starts with the package header \texttt{name\{slug\}} and then concatenates
retained files as \texttt{<file\_sep>path} followed by the file content, preserving
package boundaries and file paths without reference detection or skill-specific
reordering. Random File Order (Random File) uses the same cleaned files as Reference Insert but
independently shuffles files within each package, then shuffles serialized
skills into packs of up to 12k characters or five skills separated by
\texttt{<skill\_pack\_sep>}; this preserves package membership but removes
local file adjacency. Metadata Packing (Metadata)~\cite{gao2025metadata}
keeps the package file order and adds \texttt{<skill\_meta>} fields for id,
skill name, slug, source, and file count, as well as \texttt{<file\_meta>}
fields for index, path, inferred role, extension, and character count before
each file. Skill fill-in-the-middle (Skill-FIM)~\cite{bavarian2022efficient}
first builds the Reference
Insert serialization and, with probability 0.5, moves one semantic span of
80--2,400 characters into a middle slot marked by \texttt{<fim\_prefix>},
\texttt{<fim\_suffix>}, \texttt{<fim\_middle>}, and \texttt{<fim\_end>};
the candidate-span selection procedure prioritizes workflow, protocol, API, resource, reference,
script, and template sections before generic paragraphs.
Reference Insert is the default strategy: it starts from
\texttt{SKILL.md} when present, scans it line by line, inserts explicitly
referenced support files immediately after the referring line with
\texttt{<referenced\_file\_sep>}, appends unreferenced support files with
\texttt{<unreferenced\_file\_sep>}, and falls back to README-first package
order when no \texttt{SKILL.md} exists.

\paragraph{Reference resolution rule.}
For each cleaned package, Reference Insert first normalizes file paths by
lowercasing and replacing backslashes with slashes. Candidate identifiers
include the relative path with optional leading \texttt{./} or \texttt{/},
the basename when it has at least five characters, and the final two path
components. For every primary-file line, all matching support files not
previously inserted are ordered deterministically by extension and path and
inserted after that line. Each support file is inserted at most once, at its
first matching line. After the primary-file scan finishes, remaining support
files are appended in the package order retained by preprocessing. This rule
leaves file contents unchanged while placing referenced resources near their
first matching mentions.

Reference Insert resolves and inserts 70,655 referenced files across 17,957
packages. For each resolved reference edge, we measure the absolute
MeCo-tokenizer distance from the end of the referring line to the first token
of the target file content. Relative to serialization in the original cleaned
file order, Reference Insert reduces the mean distance from 14,518.996 to
737.189 tokens, an average reduction of 13,781.807 tokens (94.92\%).

Tables~\ref{tab:full_skill_data_organization_agentic} and
\ref{tab:full_skill_data_organization_general} report the full ablation
supporting the main-paper results. We retain individual benchmark scores because
effects vary across benchmarks and evaluation groups.

\begin{table}[!htbp]
\centering
\small
\setlength{\tabcolsep}{2pt}
\begin{tabular}{lcccc}
\toprule
\textbf{Method} & \textbf{API-Bank} & \textbf{MetaTool} & \textbf{APTBench} & \textbf{ToolEyes} \\
\midrule
DeepSeek-Coder & \underline{24.40} & 15.22 & \underline{16.14} & 25.01 \\
Random File & 17.76 & 10.55 & 14.72 & 21.34 \\
Metadata & 11.46 & 15.03 & 13.64 & 25.86 \\
Skill-FIM & 15.43 & \underline{17.06} & 13.37 & \underline{26.06} \\
Reference Insert & \textbf{26.47} & \textbf{17.39} & \textbf{17.97} & \textbf{26.39} \\
\bottomrule
\end{tabular}
\caption{
Agentic benchmark results for different serialization strategies.
}
\label{tab:full_skill_data_organization_agentic}
\end{table}

\begin{table}[!htbp]
\centering
\small
\setlength{\tabcolsep}{1mm}
\begin{tabular}{lcccccc}
\toprule
\textbf{Method} & \textbf{ARC} & \textbf{BQ} & \textbf{HS} & \textbf{PIQA} & \textbf{WG} & \textbf{MMLU} \\
\midrule
DeepSeek-Coder & 43.34 & 62.54 & \textbf{67.00} & \textbf{73.90} & 62.35 & 34.65 \\
Random File & \textbf{44.03} & 61.50 & 66.20 & 73.50 & 63.46 & \underline{35.36} \\
Metadata & 43.00 & 62.00 & \underline{66.30} & 73.50 & \textbf{64.01} & 35.30 \\
Skill-FIM & 43.60 & \underline{63.40} & 66.10 & 73.60 & 62.98 & 34.30 \\
Reference Insert & \underline{43.94} & \textbf{65.30} & 66.20 & \underline{73.80} & \underline{63.69} & \textbf{36.24} \\
\bottomrule
\end{tabular}
\caption{
General benchmark results for different serialization strategies.
}
\label{tab:full_skill_data_organization_general}
\end{table}

\section{Details of Reinforcement Learning from Verifiable Rewards (RLVR)}
\label{app:rlvr_details}

The RLVR experiments use the
SmolLM2-360M backbone~\cite{allal2025smollm2}, Tulu 3 SFT~\cite{lambert2024tulu},
and Group Relative Policy Optimization (GRPO) inspired by
DeepSeekMath~\cite{shao2024deepseekmath}, implemented with Transformers
Reinforcement Learning (TRL)~\cite{vonwerra2020trl}.

\begin{table}[!htbp]
\centering
\footnotesize
\setlength{\tabcolsep}{2pt}
\begin{tabular}{p{0.85in}p{0.90in}p{0.90in}}
\toprule
\textbf{Parameter} & \textbf{Mid-training} & \textbf{Tulu 3 SFT} \\
\midrule
GPUs & 4 $\times$ A100 80GB & 4 $\times$ A100 80GB \\
Backbone / initialization & SmolLM2-360M & Mid-training checkpoint; the no-mid-training condition uses the backbone directly \\
Training objective & Causal language modeling on fixed token blocks & Assistant-only supervised chat loss \\
Training data & Mid-training corpus (experiment-specific) & Tulu 3 \\
Total budget & 1 epoch & 3 epochs \\
Sequence length & 4,096 & 2,048 \\
Per-device batch size & 1 & 1 \\
Gradient accum.\ steps & 8 & 16 \\
Learning rate & $2\times10^{-5}$ & $2\times10^{-5}$ \\
Weight decay & 0.1 & 0.0 \\
Optimizer & Adafactor & Adafactor \\
Scheduler & cosine & cosine \\
Precision & BF16; TF32 enabled & BF16; TF32 enabled \\
Gradient checkpointing & enabled & enabled \\
No. of seeds & 5 & 5 \\
\bottomrule
\end{tabular}
\caption{
Mid-training and SFT hyperparameters for the SmolLM2-360M RLVR experiments.
}
\label{tab:rlvr_cpt_sft_hyperparameters}
\end{table}

\begin{table}[!htbp]
\centering
\footnotesize
\setlength{\tabcolsep}{2pt}
\begin{tabular}{p{0.85in}p{1.90in}}
\toprule
\textbf{Parameter} & \textbf{RLVR (GRPO)} \\
\midrule
GPUs & 4 $\times$ A100 80GB \\
Backbone / initialization & SFT checkpoint (per group) \\
Training objective & GRPO with verifiable rewards \\
Training data & RLVR-GSM-MATH-IF-Mixed-Constraints \\
Total budget & 1 epoch \\
Prompt/completion lengths & 1,024 / 256 \\
Per-device batch size & 2 \\
Gradient accum.\ steps & 4 \\
Learning rate & $1\times10^{-6}$ \\
Kullback--Leibler (KL) weight $\beta$ & 0.04 \\
No. of generations & 4 \\
Optimizer & Adafactor \\
Scheduler & cosine, 3\% warmup \\
Precision & BF16 \\
Gradient checkpointing & enabled \\
No. of seeds & 5 \\
\bottomrule
\end{tabular}
\caption{
RLVR (GRPO) hyperparameters for the SmolLM2-360M experiments.
All evaluated configurations (No mid-training, Dolmino, \textsc{SkillCorpus}) share identical RLVR settings.
}
\label{tab:rlvr_hyperparameters}
\end{table}

Tables~\ref{tab:full_rlvr_agentic} and \ref{tab:full_rlvr_general} give the
full RLVR benchmark scores behind the main-paper summary.

\begin{table}[!htbp]
\centering
\small
\setlength{\tabcolsep}{2pt}
\begin{tabular}{lcccc}
\toprule
\textbf{Mid-training} & \textbf{API-Bank} & \textbf{MetaTool} & \textbf{APTBench} & \textbf{ToolEyes} \\
\midrule
No mid-training & 5.44 & 1.16 & 4.54 & 8.64 \\
Dolmino & \underline{7.69} & \underline{4.75} & \underline{7.66} & \underline{8.90} \\
\textsc{SkillCorpus} & \textbf{15.93} & \textbf{10.88} & \textbf{9.43} & \textbf{11.52} \\
\bottomrule
\end{tabular}
\caption{
Full agentic benchmark results after RLVR post-training.
All models use the SmolLM2-360M backbone.
}
\label{tab:full_rlvr_agentic}
\end{table}

\begin{table}[!htbp]
\centering
\small
\setlength{\tabcolsep}{1mm}
\begin{tabular}{lcccccc}
\toprule
\textbf{Mid-training} & \textbf{ARC} & \textbf{BQ} & \textbf{HS} & \textbf{PIQA} & \textbf{WG} & \textbf{MMLU} \\
\midrule
No mid-training & 23.21 & 62.80 & 24.10 & 49.20 & 49.70 & 31.57 \\
Dolmino & \underline{24.66} & \underline{63.50} & \underline{24.60} & \underline{50.20} & \underline{50.43} & \underline{32.36} \\
\textsc{SkillCorpus} & \textbf{24.83} & \textbf{63.80} & \textbf{26.00} & \textbf{51.20} & \textbf{50.67} & \textbf{32.43} \\
\bottomrule
\end{tabular}
\caption{
Full general benchmark results after RLVR post-training.
}
\label{tab:full_rlvr_general}
\end{table}

\section{Artifact Availability}

We will publicly release the complete corpus-construction, cleaning,
decontamination, and Reference Insert code; all training and evaluation
configurations; and all mid-training, post-training, and final checkpoints
corresponding to the experimental conditions reported in this work.

\bibliography{custom}

\end{document}